\pdfoutput=1
\documentclass[11pt,dvipsnames]{article}

\usepackage{emnlp2021}

\usepackage{times}
\usepackage{latexsym}
\usepackage{comment}

\usepackage[T1]{fontenc}
\usepackage[utf8]{inputenc}

\usepackage{microtype}

\usepackage{booktabs}
\usepackage{amsmath}
\usepackage{amsfonts}
\usepackage{graphicx}
\usepackage{subcaption}
\usepackage{multirow}
\usepackage{pifont}% http://ctan.org/pkg/pifont
\title{On Releasing Annotator-Level Labels and Information in Datasets}

\author{Vinodkumar Prabhakaran\thanks{\hspace{.4em}Authors contributed equally} \\
  Google Research \\
  \texttt{vinodkpg@google.com} \\\And
  Aida Mostafazadeh Davani\footnotemark[1]\hspace{.1em} \thanks{\hspace{.4em}Work done while at Google Research} \\
  University of Southern California \\
  \texttt{mostafaz@usc.edu} \\\AND
  Mark D\'iaz \\
  Google Research \\
  \texttt{markdiaz@google.com} \\}

\begin{document}
\maketitle
\begin{abstract}

A common practice in building NLP datasets, especially using crowd-sourced annotations, involves obtaining multiple annotator judgements on the same data instances, which are then flattened to produce a single ``ground truth'' label or score, through majority voting, averaging, or adjudication. While these approaches may be appropriate in certain annotation tasks, such aggregations overlook the socially constructed nature of human perceptions that annotations for relatively more subjective tasks are meant to capture. 
% In particular, annotators' socio-cultural backgrounds and lived experiences may cause them to disagree systematically with each other. 
In particular, systematic disagreements between annotators owing to their socio-cultural backgrounds and/or lived experiences are often obfuscated through such aggregations. 
In this paper, we empirically demonstrate that label aggregation may introduce representational biases of individual and group perspectives. Based on this finding, we propose a set of recommendations for increased utility and transparency of datasets for downstream use cases.

\end{abstract}

\section{Introduction}

% \vp{Edit this section heavily, as it is copied over from our TACL intro.}
Obtaining multiple annotator judgements on the same data instances is a common practice in NLP in order to improve the quality of final labels \cite{snow2008cheap,nowak2010reliable}.
% snow-etal-2008-cheap,
% Annotation practices %, especially through crowd-sourcing
% often obtain multiple annotator judgements on each data instance to improve the quality of final labels \cite{snow-etal-2008-cheap,nowak2010reliable}. 
Cases of disagreement between annotations are often resolved through majority voting, averaging, or adjudication in order to derive a single  ``ground truth'', often with the aim of training supervised machine learning models. 
% \cite{sabou2014corpus}
% Annotation aggregation essentially derives a singular ground truth or gold label for further use in supervised model training.
However, in relatively subjective tasks such as sentiment analysis or offensiveness detection, there often exists no single ``right'' answer \citep{alm2011subjective}. Enforcing such a single ground truth in such tasks will sacrifice valuable nuances about the task that are embedded in annotators' assessments of the stimuli,  especially their disagreements \cite{aroyo2013crowd}.

Annotators' socio-demographic factors, moral values, and lived experiences often influence their interpretations of language, especially in subjective tasks such as identifying political stances \citep{luo-etal-2020-detecting}, sentiment \cite{diaz2018addressing}, and online abuse \cite{waseem2016you,patton2019annotating}.
% cowan2003empathy
% Such differences in perceptions will result in systematic disagreement between annotators, which is lost during aggregation. Furthermore, majority vote may result in ignoring minority perspectives in data, further amplifying societal biases present in data. 
% 
% 
% 
% 
% Insights from disagreements are crucial for modeling and investigating subjective tasks, such as identifying political stances, online abuse, and hate speech, attitudes toward which tend to differ across demographic groups \cite{cowan2003empathy}.
For instance, feminist and anti-racist activists systematically disagree with crowd workers in their hate speech annotations \cite{waseem2016you}. 
Similarly, annotators' political affiliation is shown to correlate with how they annotate the neutrality of political stances \citep{luo-etal-2020-detecting}. 
A potential adverse effect of majority voting in such cases is that it may sideline minority perspectives in data.

In this paper, we analyze annotated data for eight different tasks across three different datasets to study the impact of majority voting as an aggregation approach. We answer two questions: 

\begin{itemize}
\item \textbf{Q1}: Does aggregated data uniformly capture all annotators' perspectives, when available? 
% Or, does it reflect certain annotators' perspectives more than others?
\item \textbf{Q2}: Does aggregated data reflect certain socio-demographic groups' perspectives more so than others? 
\end{itemize}

\noindent Our analysis demonstrates that in the annotations for many tasks, the aggregated majority vote does not uniformly reflect the perspectives of all annotators in the annotator pool. 
% While for some tasks, the majority vote labels did have about the same agreement scores with all individual annotators, f
For many tasks in our analysis, a significant proportion of the annotators had very low agreement scores (0 to 0.4) with the majority vote label. While certain individual annotator's labels may have low agreement with the majority label due to valid/expected reasons (e.g., if they produced noisy labels), we further show that these agreement scores may vary significantly across different socio-demographic groups that annotators identify with. This finding has important fairness implications, as it demonstrates how the aggregation step can sometimes cause the final dataset to under-represent certain groups' perspectives.  
% suggests that the aggregation step may cause the final dataset to under-represent certain groups' perspectives. 

Meaningfully addressing such issues in multiply-annotated datasets requires understanding and accounting for systematic disagreements between annotators. However, most annotated datasets often only release the aggregated labels, without any annotator-level information. 
We argue that dataset developers should consider including annotator-level labels as well as annotators' socio-demographic information (when viable to do so responsibly) when releasing datasets, especially those capturing relatively subjective tasks.
Inclusion of this information will enable more research on how to account for systematic disagreements between annotators in training tasks.

\begin{comment}
In this paper, we argue in favor of including annotator-level labels as well as annotators' socio-demographic information when releasing datasets.
% , as a way to mitigate some of these issues. 
Inclusion of this information will enable more research on how to account for systematic disagreements between annotators in downstream training tasks. To support our argument we analyze the impact of majority voting as an aggregation approach in annotated data for eight different tasks across three different datasets. We ask two questions:
\begin{itemize}
\item \textbf{Q1}: Does aggregated data uniformly capture all annotators' perspective, when available? Or, does it reflect certain annotators' perspectives more than others?
% \item \textbf{Q1}: Does majority voting disregard some annotators entirely, consequently excluding them from the final dataset?
\item \textbf{Q2}: Does aggregated data reflect certain socio-demographic groups' perspectives more favorably than others? 
% \item \textbf{Q2}: Are under-represented annotators from specific (possibly minority) social groups?
\end{itemize}
If the answer to \textbf{Q1} and \textbf{Q2} are both positive, we conclude that not only majority voting is not a justifiable approach for dealing with several annotations and their possible disagreement, but also, socio-demographic information about annotators should be provided for data users in order to identify the effects of annotator differences on subjective tasks and mitigate the impact of imbalances datasets.
\end{comment}

% \begin{comment}

% \end{comment}

\section{Background}
\label{sec:approaches}

NLP has a long history of developing techniques to interpret subjective language \cite{wiebe2004learning,alm2011subjective}. While all human judgments embed some degree of subjectivity, some tasks such as sentiment analysis \cite{liu2010sentiment}, affect modeling \cite{alm2008affect,liu2003model}, emotion detection \cite{hirschberg2003experiments}, and hate speech detection \cite{warner2012detecting} are agreed upon as relatively more subjective in nature. 
As \citet{alm2011subjective} points out, achieving a single \textit{real `ground truth'} is not possible, nor essential in case of such subjective tasks. Instead, we should investigate how to model the subjective interpretations of the annotators, and how to account for them in application scenarios.

However, the current practice in the NLP community continues to be applying different aggregation strategies to arrive at a single score or label that makes it amenable to train and evaluate supervised machine learning models. Oftentimes, datasets are released with only the final scores/labels, essentially obfuscating important nuances in the task. The information released about the annotations 
can be at one of the following four levels of information-richness. 
% can be thought of across the following four levels of information-richness (with increasing degree of information).
% from just the aggregated labels (most common approach) to richer annotator-level labels and information, which is relatively rare:

% \paragraph{Aggregated labels:} This is the most common approach, 
Firstly, the most common approach is
one in which multiple annotations obtained for a data instance are aggregated to derive a single ``ground truth'' label, and these labels are the only annotations included in the released dataset (e.g., \citet{founta2018large}). The aggregation strategy most commonly used, especially in large datasets, is \textit{majority voting}, although smaller datasets sometimes use \textit{adjudication} by an `expert' (often one of the study authors themselves) to arrive at a single label (e.g., in \citet{waseem2016hateful}) when there are substantial disagreements between annotators. These aggregation approaches rely on the assumption that there always exist a single correct label, and that either the majority label or the `expert' label is more likely to be that correct label. What it fails to account for is the fact that in many subjective tasks, e.g., detecting hate speech, the perceptions of individual annotators may be as valuable as an `expert' perspective.

% \paragraph{Soft labels:} 
Secondly, some datasets (e.g., \citet{Jigsaw-toxic,davidson2017automated}) release the distribution across labels rather than a single aggregated label. In binary classification tasks, this corresponds to the percentage of annotators who chose one of the labels. In multi-class classification, this may be the distribution across labels obtained for an instance. 
% More advanced aggregation methods may employ a weighted average that takes into account the reliability or trustworthiness of individual annotators \citep{fornaciari-etal-2021-beyond}. 
While this provides more information than a single aggregated label does (e.g., identifies the instances with high disagreement), it fails to capture annotator-level systematic differences.

% \paragraph{Annotator-level labels:} 
Thirdly, some datasets release annotations made by each individual annotators in an anonymous fashion (e.g., \citet{kennedy2020gab,Jigsaw-bias}). Such annotator-level labels allow downstream dataset users to investigate and account for systematic differences between individual annotators' perspectives on the tasks, although they do not contain any information about each annotators' socio-cultural backgrounds.
% 
% \paragraph{Annotator-level labels and socio-demographic information:} 
Finally, some recent datasets (e.g., \citet{diaz2018addressing}) also release such socio-demographic information about the annotators in addition to annotator-level labels. This information may include various identity subgroups the annotators self-identify with (e.g., gender, race, age range, etc.), or survey responses from the annotators that capture their value systems, lived experiences, or expertise, as they relate to the specific task at hand. Such information, while tricky to share responsibly, would help enable analysis around representation of marginalized perspectives in datasets, as we demonstrate in the next section.

\begin{figure*}
\centering
    % \begin{subfigure}[b]{.75\textwidth}
        \includegraphics[width=0.29\textwidth]{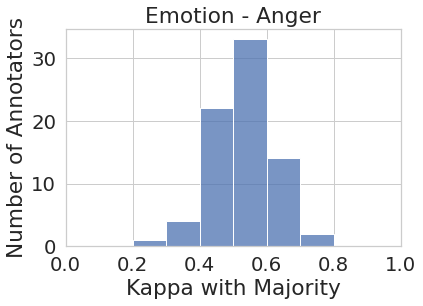}
        \includegraphics[width=0.29\textwidth]{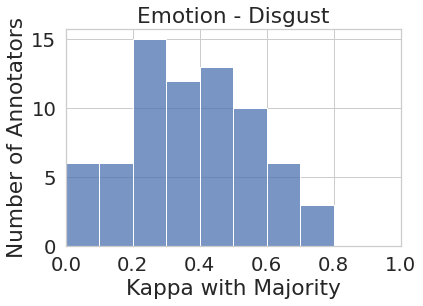}
        \includegraphics[width=0.29\textwidth]{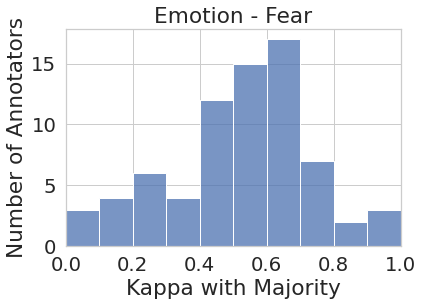}\\\vspace{2mm}
        \includegraphics[width=0.29\textwidth]{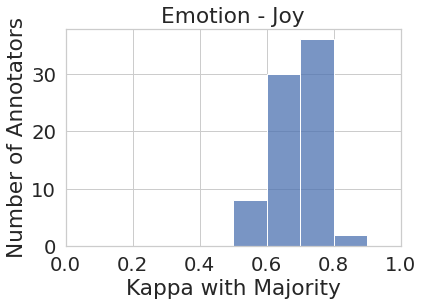}
        \includegraphics[width=0.29\textwidth]{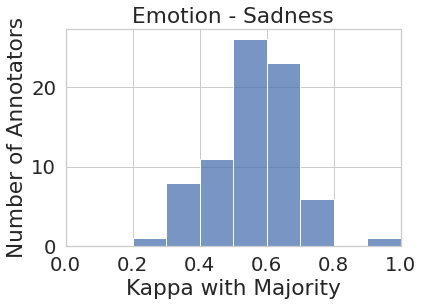}
        \includegraphics[width=0.29\textwidth]{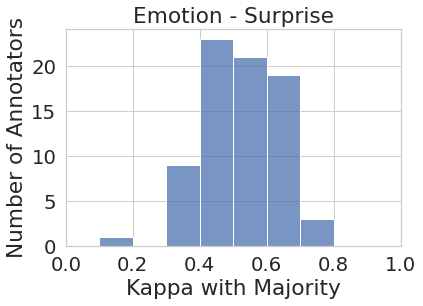}\\\vspace{2mm}
        % \caption{GoEmotions}
    % \end{subfigure}
    % \begin{subfigure}[b]{.24\textwidth}
        \includegraphics[width=0.29\textwidth]{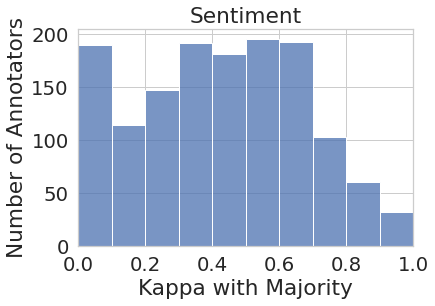}
        \includegraphics[width=0.29\textwidth]{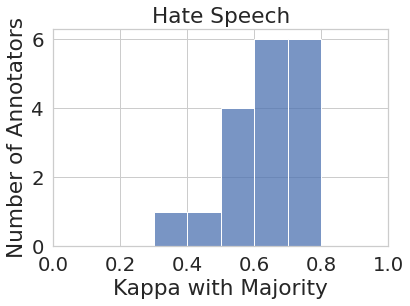}
        % \caption{Sentiment and Hate Speech}
    % \end{subfigure}
    \caption{Histograms represent the frequency distribution of annotator agreement with the aggregated label for eight tasks under three datasets for Emotions, Sentiment and Hate Speech datasets. The lack of uniformity in the distributions means that annotator perspectives are not equally captured in the majority labels.}
    \label{fig:majority_kappa}
\end{figure*}

\section{Impacts of Aggregation}
% \section{Impacts of Aggregation on Representation}
\label{sec:analyses}

In this section, we investigate how the aggregation of multiple annotations to a single label impact representations of individual and group perspectives in the resulting datasets. We analyze annotations for eight binary classification tasks, across three different datasets: hate-speech \cite{kennedy2020gab},  sentiment \cite{diaz2018addressing}, and emotion \cite{demszky2020goemotions}. Table~\ref{tab:datasets} shows the number of instances, annotators and individual annotations present in the datasets. For hate-speech and emotion datasets, we use the binary label in the raw annotations, whereas for the sentiment dataset, we map the 5-point ordinal labels (-2, -1, 0, +1, +2) in the raw data to a binary distinction denoting whether the text was deemed positive or negative.\footnote{We do this mapping for the purposes of this analysis, where we are focusing on binary tasks. Ideally, a more nuanced 5-point labeling schema will be more useful.} While the emotion dataset contains annotations for 28 different emotions, in this work, for brevity, we focused on the annotations for only the six standard Ekman emotions \cite{ekman1992argument} --- \textit{anger}, \textit{disgust}, \textit{fear}, \textit{joy}, \textit{sadness}, and \textit{surprise}. In particular, we use the raw annotations for these six emotions, rather than the mapping of all 28 emotions onto these six emotions that \citet{demszky2020goemotions} use in some of their experiments.

\begin{table}[t!]
\small
\centering
\begin{tabular}{@{}lccc@{}}
\toprule
%&&& \multicolumn{3}{c}{Released Information}\\\cline{4-6}
Dataset & \#instances & \#annotators & \#annotations \\% &  Aggregated & Annotator-level & Annotator Demographics\\ 
\midrule
% \citet{waseem2016you} & Hate-speech & \\%& & & \\
Hate-speech  & 27,665 & 18 & 86,529\\%& \xmark & \cmark & \xmark \\
% \citet{Jigsaw-bias} & Toxicity & \\%& \cmark & \cmark & \xmark \\
Sentiment  & 14,071 & 1,481 & 59,240 \\% & \cmark & \cmark & \cmark \\
% \citet{danescu2013computational} & Politeness & \\%& \cmark & \cmark & \xmark \\
Emotion & 58,011 & 82 & 211,225\\%& \cmark & \cmark & \xmark \\

\bottomrule
\end{tabular}
\caption{Statistics on the three datasets we analyze: Hate-speech \cite{kennedy2020gab},  Sentiment \cite{diaz2018addressing}, and Emotions \cite{demszky2020goemotions}}
\label{tab:datasets}
\end{table}

\begin{figure*}
    \centering
    \includegraphics[width=0.31\textwidth]{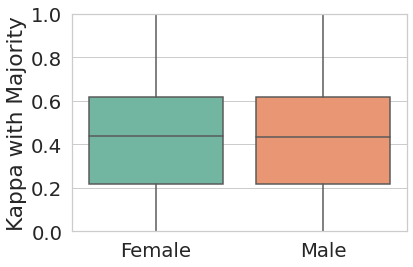}\hspace{4mm}
    \includegraphics[width=0.31\textwidth]{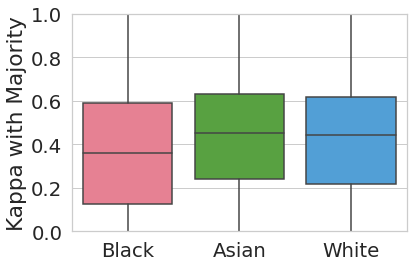}\hspace{4mm}
    \includegraphics[width=0.31\textwidth]{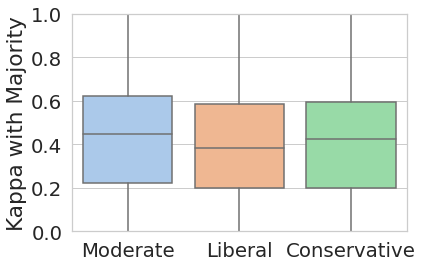}
    \caption{Average and standard deviation of annotator agreement with aggregated labels, calculated for annotators grouped by their socio-demographics under gender, race, and political affiliation.}
    \label{fig:sentiment-stat}
\end{figure*}
\subsection{Q1: Do Aggregated Labels Represent Individual Annotators Uniformly?}
First, we investigate whether the aggregated labels obtained through majority labels provide a more or less equal representation for all annotator perspectives.
% , or whether it leads to significant under-representation of certain annotators. 
For this analysis, we calculate the majority label for each instance as the label that half or more annotators who annotated that instance agreed on. We then measure Cohen's Kappa agreement score for each individual annotator's labels and the majority labels on the subset of instances they annotated. While lower agreement scores between some individual annotators and the majority vote is expected (e.g., if the annotator produced noisy labels, or they misunderstood the task), the assumption is that the majority label captures the perspective of the `average human annotator' within the annotator pool. 

Figure \ref{fig:majority_kappa} represents the histogram of annotators' agreement scores with majority votes for all eight tasks. While the majority vote in some tasks such as \textit{joy} and \textit{sadness} (to some extent) do represent most of the annotator pool more or less uniformly (i.e., majority vote agrees with most annotators at around the same rate), in most cases, the majority vote under-represents or outright ignores the perspectives of a substantial number of annotators. For instance, majority vote for \textit{disgust} has very low agreement ($\kappa < 0.3$) with almost one-third (27 out of 82) of the annotator pool. Similarly, majority vote for \textit{sentiment} has very low agreement with around one-third (450+) of their annotator pool. 

% This side effect of aggregation is consequential in what perspectives will be learned by downstream machine learning models. \vp{Add a stronger concluding sentence or two, and teeing up the next subsection}.

\subsection{Q2: Do Aggregated Labels Represent All Social Groups Uniformly?}

While the analysis on \textbf{Q1} reveals that certain annotator perspectives are more likely to be ignored in the majority vote, it is especially problematic from a fairness perspective, if these differences vary across different social groups. Here, we investigate whether specific socio-demographic groups and their perspectives are unevenly disregarded through annotation aggregation. To this end, we analyze the sentiment analysis dataset \citep{diaz2018addressing} since it includes raw annotations as well as annotators' self-identified socio-demographic information. Furthermore, as observed in Figure \ref{fig:majority_kappa}, a large subset of annotators in this dataset are in low agreement with the aggregated labels. 

We study three demographic attributes, namely race, gender, and political affiliation and compare the agreement scores between the aggregated labels and the individual annotators' labels within each group. Figure \ref{fig:sentiment-stat} shows the average and standard deviation of annotators' agreement scores with aggregated labels for each demographic group:
% of annotators with the same demographic attributes according to their 
race (\textit{Asian}, \textit{Black}, and \textit{White}), gender (\textit{Male}, and \textit{Female}), and political affiliation (\textit{Conservative}, \textit{Moderate}, and \textit{Liberal}).\footnote{We removed social groups with fewer than 50 annotators from this analysis for lack of sufficient data points. These include other racial groups such as `Middle Eastern' with 2 annotators, `Native Hawaiian or Pacific Islander' with 4 annotators, and non-binary gender identity with one annotator).} 

We perform three one-way ANOVA tests to test whether annotators belonging to different demographic groups have significantly different agreement scores with the aggregated labels, on average.
% We performed three one-way ANOVA tests to assess whether the agreement (the dependent variable) between annotators and the aggregated labels  were different across sociodemographic groups. 
The results show significant differences among racial groups ($F(2, 2387)$=3.77, $p$=0.02); in particular, White annotators show an average agreement of 0.42 ($SD$=0.26), significantly higher (%$F(1,1294)$=, 
$p$=0.03 according to a post-hoc Tukey test) than Black annotators with average of 0.37 ($SD$=0.27).
% and Asian annotators with an average of 0.44 ($SD$ =  0.26) do not show significant difference with other two groups. 
The difference between average agreement scores across different political groups are not statistically significant, although
moderate annotators on average have higher agreement (0.42) compared to conservative and liberal annotators (0.40 and 0.38, respectively).
% The average agreement scores are not significantly different among different political groups, although
% ($F(2, 1440)$=1.82, $p$=0.16), 
% as moderate annotators on average have 0.42 ($SD$=0.26) agreement with the aggregated labels and conservative and liberal annotators have an average agreement of 0.40 ($SD$=0.26) and 0.38 ($SD$=0.27) respectively. 
Similarly, annotation agreements of male and female annotators are not significantly different.

\section{Utility of Annotator-level Labels}

% \md{Just dropping an intrusive thought while I should be working on the HCOMP paper ;) that it would be worth discussing that developing models to predict or measure subjective phenomena should involve explicit consideration of annotator social context/background rather than starting from the assumption that annotators are interchangeable until proven to be otherwise.}

Another argument in favor of retaining annotator-level labels is their utility in modeling disagreement during training and evaluation. 
% Annotator-level labels has been shown to be useful during model training. 
\citet{prabhakaran2012statistical} and 
\newcite{plank2014learning} incorporated annotator disagreement in the loss functions used during training to improve predictive performance. 
\citet{cohn2013modelling} and \citet{fornaciari-etal-2021-beyond} use a multi-task approach to incorporate annotator disagreements to improve machine translation and part-of-speech tagging performance, respectively.
\citet{chou2019every} and \citet{guan2018said} developed learning architectures that model individual annotators as a way to improved performance. \citet{wich2020investigating} show the utility of detecting clusters of annotators in hate-speech detection based on how often they agree with each other. Finally, \citet{davani2021dealing} introduce a multi-annotator architecture  that models each annotators' perspectives separately using a multi-task approach. They demonstrate that this architecture helps to model uncertainty in predictions, without any significant loss of accuracy or efficiency.
% annotator-level labels help model uncertainty in predictions.
This array of recent work further demonstrates the utility of retaining annotator-level information in the datasets for downstream modeling steps.

\section{Discussion and Conclusion}
% balayn2018characterising
Building models to predict or measure subjective phenomena based on human annotations should involve explicit consideration for the unique perspectives each annotator brings forth in their annotations. Annotators are not interchangeable-- that is, they draw from their socially-embedded experiences and knowledge when making annotation judgments. As a result, retaining their perspectives separately in the datasets will enable dataset users to account for these differences according to their needs.
% Our two analyses, described in Section~\ref{sec:analyses}, reveals
We demonstrated that annotation aggregation may unfairly disregard perspectives of certain annotators, and sometimes certain socio-demographic groups. 
% Moreover, releasing the aggregated labels rather than annotator-level labels prevents alternative practices for dealing with the impact of aggregation on representation. 
Based on our analysis, we propose three recommendations aimed to avoid these issues:
% avoid introducing representational biases in the dataset construction stage: 
% potentially resolve this shortcoming.

% \vp{Edit heavily}
% \vspace{-2pt}
\paragraph{Annotator-level labels:}
%As discussed in Sec.~\ref{sec:approaches} most multiply annotated datasets are released with no annotator-level information, and instances are labeled by the aggregated votes \cite{waseem2016hateful, Jigsaw-toxic}, or aggregate percentages \cite{davidson2017automated,Jigsaw-bias}. As an alternative, dataset curation practices can release 
We urge dataset developers to release the annotator-level labels, preferably in an anonymous fashion, and leave open the choice of whether and how to utilize or aggregate these labels for the dataset users.

% \vspace{-2pt}
\paragraph{Socio-demographic information:} 
% Annotators' sociodemographic identity may significantly shape their perspectives on subjective tasks. Hence, the
Sociodemographic identity of the annotators is crucial to ascertain that the datasets (and the models trained on them) equitably represent perspectives of various social groups. We urge dataset developers to include socio-demographic information of annotators, when viable to do so responsibly.

% \vspace{-2pt}
\paragraph{Documentation about recruitment, selection, and assignment of annotators:}
Finally, we urge dataset developers to document how the annotators were recruited, the criteria used to select them and assign data to them, and any efforts to ensure representational diversity, through transparency artefacts such datasheets \cite{gebru2018datasheets} or data statements \cite{bender2018data}.

\section*{Acknowledgements}

We thank Ben Hutchinson for valuable feedback on the manuscript. We also thank the anonymous reviewers for their feedback.

% \section{Conclusion}

%clearpage

\bibliography{anthology,custom}
\bibliographystyle{acl_natbib}

% \appendix

% \section{Example Appendix}
% \label{sec:appendix}

% This is an appendix.

\end{document}